\documentclass[10pt,twocolumn,letterpaper]{article}

\usepackage{iccv}
\usepackage{times}
\usepackage{epsfig}
\usepackage{graphicx}
\usepackage{amsmath}
\usepackage{amssymb}
\usepackage{amsthm}
\usepackage{mathrsfs}
\usepackage{float}
\usepackage{array}
\usepackage{multirow}
\usepackage{longtable}
\usepackage{color}
\usepackage{caption}
\usepackage{booktabs}

\usepackage[pagebackref=true,breaklinks=true,letterpaper=true,colorlinks,bookmarks=false]{hyperref}

\iccvfinalcopy 

\def\iccvPaperID{3372} 

\ificcvfinal\fi

\begin{document}

\title{ReCU: Reviving the Dead Weights in Binary Neural Networks}

\author{Zihan Xu$^1$,\;\; Mingbao Lin$^1$,\;\; Jianzhuang Liu$^3$,\;\; Jie Chen$^{4,5}$ \\ Ling Shao$^6$,\;\; Yue Gao$^7$,\;\; Yonghong Tian$^{4,5}$,\;\; Rongrong Ji$^{1,2,5}$\thanks{Corresponding author: rrji@xmu.edu.cn} \\
$^1$MAC Lab, School of Informatics, Xiamen University\\
$^2$Institute of Artificial Intelligence, Xiamen University\\
$^3$Noah's Ark Lab, Huawei Technologies \\ $^4$School of Electronic and Computer Engineering, Peking University \\ $^5$Peng Cheng Lab \;\;\;\; $^6$Inception Institute of Artificial Intelligence \\
$^7$School of Software, THUIBCS, BNRist, Tsinghua University \\
}

\maketitle
\ificcvfinal\thispagestyle{empty}\fi

\begin{abstract}
Binary neural networks (BNNs) have received increasing attention due to their superior reductions of computation and memory. Most existing works focus on either lessening the quantization error by minimizing the gap between the full-precision weights and their binarization or designing a gradient approximation to mitigate the gradient mismatch, while leaving the ``dead weights'' untouched. This leads to slow convergence when training BNNs. In this paper, for the first time, we explore the influence of ``dead weights'' which refer to a group of weights that are barely updated during the training of BNNs, and then introduce rectified clamp unit (ReCU) to revive the ``dead weights'' for updating. We prove that reviving the ``dead weights'' by ReCU can result in a smaller quantization error. Besides, we also take into account the information entropy of the weights, and then mathematically analyze why the weight standardization can benefit BNNs. We demonstrate the inherent contradiction between minimizing the quantization error and maximizing the information entropy, and then propose an adaptive exponential scheduler to identify the range of the ``dead weights''. By considering the ``dead weights'', our method offers not only faster BNN training, but also state-of-the-art performance on CIFAR-10 and ImageNet, compared with recent methods. Code can be available at \url{https://github.com/z-hXu/ReCU}.
\end{abstract}

\section{Introduction}

Deep Neural Networks (DNNs) have shown tremendous success and advanced many visual tasks~\cite{Krizhevsky2017ImageNetCW, Redmon2016YouOL, Everingham2009ThePV, simonyan2014very}. Nevertheless, this comes at the price of massive memory usage and computational burden, which poses a great challenge to the resource-constrained cutting-edge devices such as mobile phones and embedded devices. The community has proposed various approaches to solve this problem. Typical techniques include, but are not limited to, efficient architecture design~\cite{Han2020GhostNetMF, howard2017mobilenets, Ma2018ShuffleNetVP}, knowledge distillation~\cite{Hinton2015DistillingTK, Kim2018ParaphrasingCN, romero2014fitnets}, network pruning~\cite{Ding2019GlobalSM,  Lin2017RuntimeNP, Liu2019MetaPruningML}, and network quantization~\cite{Zhou2016DoReFaNetTL,Zhuang2018TowardsEL, Yang2020SearchingFL, ajanthan2021mirror}.

\begin{figure}[t]
\begin{center}
\includegraphics[width=0.6\linewidth]{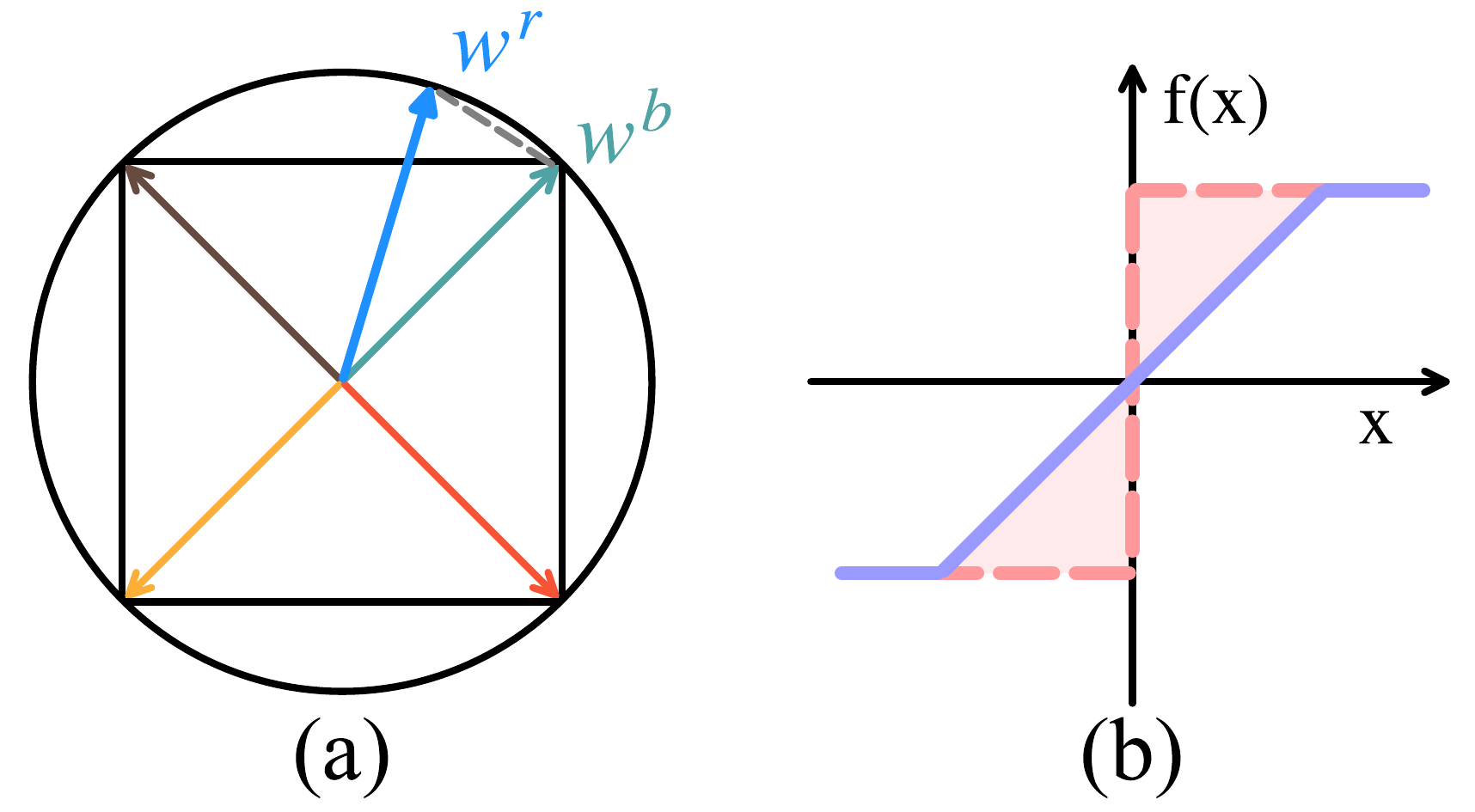}
\end{center}
\vspace{-1.5em}
   \caption{Illustration of the quantization error (a) and the gradient mismatch (b).}
\vspace{-1.5em}
\label{Fig:fig1}
\end{figure}

Among them, by converting the full-precision parameters and activations into low-bit forms, network quantization has offered a promising solution to yield a light and efficient version of DNNs~\cite{Jacob2018Quantization, Gong2019DifferentiableSQ, Zhang2018LQNetsLQ, Yang2019QuantizationN}. In the extreme case of a 1-bit representation, a binary neural network (BNN) restricts the weights and activations to only two possible values, \emph{i.e.}, -1 and +1. In comparison with the original networks, BNNs show overwhelming superiority in reducing the model complexity by around 32$\times$ parameter compression, and 58$\times$ speedup, using the efficient XNOR and bitcount operations~\cite{Bulat2019XNORNetIB}.

Despite the superiority of BNNs in memory saving and computation reduction, they suffer a drastic drop in accuracy compared with their real-valued counterparts~\cite{Rastegari2016XNORNetIC, courbariaux2016binarized, Courbariaux2015BinaryConnectTD}, which greatly limits the practical deployment. There are two main reasons for the performance degradation: large quantization error in the forward propagation and gradient mismatch during backpropagation.

Specifically, quantization error refers to the residual between the full-precision weight vector and its binarization~\cite{Rastegari2016XNORNetIC,Lin2020RotatedBN}, as illustrated in Fig.\,\ref{Fig:fig1}(a). The representational ability of BNNs is indeed limited upon the vertices of a unit square. In contrast, the full-precision weights possess an almost unlimited representation space. Such a representation gap easily results in a large accumulated error when mapping the real-valued weights into the binary space. To solve this, existing approaches try to lessen the quantization error by introducing a scaling factor to reduce the norm difference~\cite{Bulat2019XNORNetIB,Bulat2019ImprovedTO}, or devising a rotation matrix to align the angle bias~\cite{Lin2020RotatedBN}. Gradient mismatch comes from the disagreement between the presumed and actual gradient functions~\cite{lin2016overcoming} as illustrated in the pink area of Fig.\,\ref{Fig:fig1}(b). Since the quantization function in the forward propagation of BNNs has zero gradient almost everywhere, an approximate gradient function is required to enable the network to update. A typical example is the straight through estimator (STE)~\cite{Bengio2013EstimatingOP}, which however leads to inaccurate optimization directions and thus hurts the stability of network training, especially in the low bitwidth~\cite{Alizadeh2019AnES, Bulat2020BATSBA}. To mitigate this, a large collection of works have been proposed, typically by adjusting the network structures~\cite{liu2018bi,liu2020reactnet,Chen2020BinarizedNA, bulat2020high}, or using gradient functions that gradually approach zero~\cite{Gong2019DifferentiableSQ,Yang2019QuantizationN,Lin2020RotatedBN}.

\begin{figure}[!t]
\begin{center}
\includegraphics[width=0.7\linewidth]{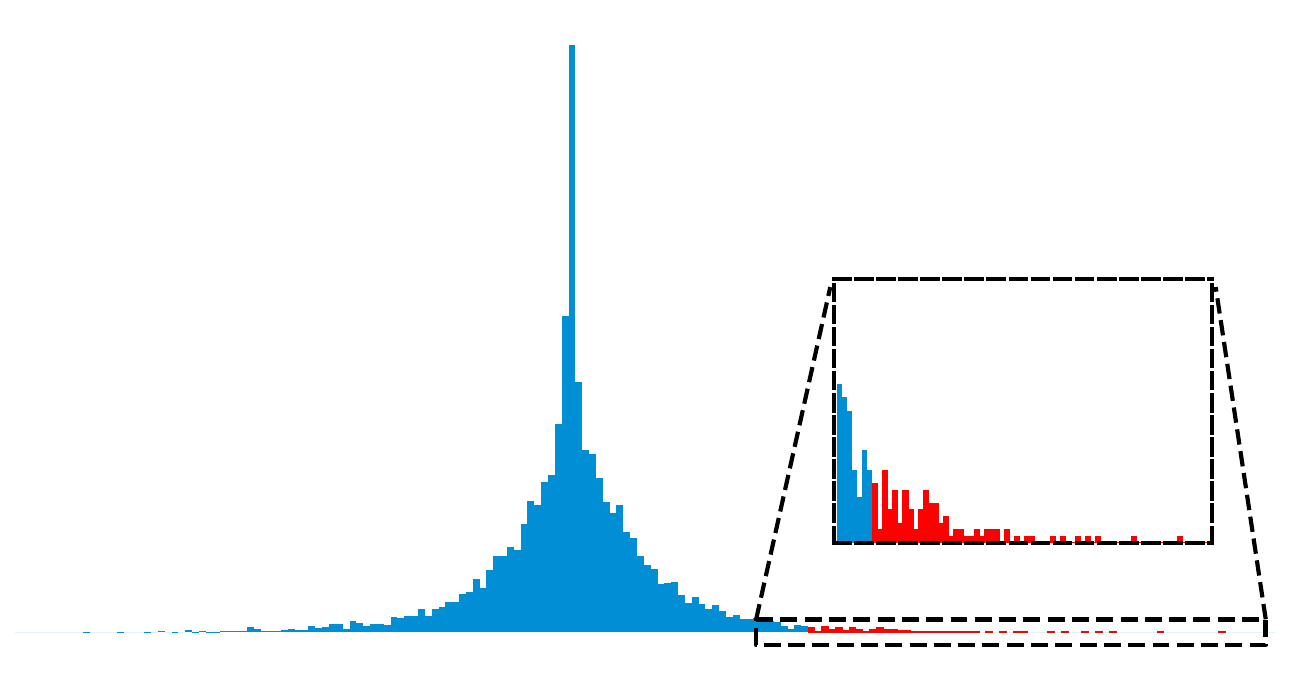}
\end{center}
\vspace{-1em}
   \caption{Illustration of the ``dead weights''. The red outliers in both tails of the distribution are barely updated during training BNNs and affect the convergence of the training (Layer2.0.conv2 of ResNet-20). (Best viewed in color)}
\vspace{-1em}
\label{dead_weight_illu}
\end{figure}

In this paper, we present a novel perspective to improve the effectiveness and training efficiency of BNNs. Inspired by~\cite{Helwegen2019LatentWD}, the latent weights, which refer to the real-valued weights used during backpropagation, play an important role in binarizing DNNs. We explore the real-valued weights of a given DNN and find that the weights falling into the two tails of the distribution, as shown in Fig.\,\ref{dead_weight_illu}, are barely updated during the training of BNNs. We call them ``dead weights'' and find that they harm the optimization and slow down the training convergence of BNNs. To solve this problem, we develop a rectified clamp unit (ReCU), which aims to revive the ``dead weights'' by moving them towards the distribution peak in order to increase the probability of updating these weights. Through a rigorous analysis, we demonstrate that the quantization error after applying ReCU is a convex function, and thus can be further reduced. Instead of simply minimizing the quantization error, we consider the information entropy of the weights to increase the weight diversity of BNNs. For the first time, a systematical analysis is derived to explain why the weight standardization~\cite{Qin2020ForwardAB} can boost the performance of BNNs, and then a generalized weight standardization is proposed to further increase the information entropy. Combining the information entropy and the quantization error, we reveal the inherent contradiction between maximizing the former and minimizing the latter, and then propose an adaptive exponential scheduler to identify the range of the ``dead weights'' and balance the information entropy of the weights and the quantization error.

We conduct extensive experiments for binarizing networks including ResNet-18/20~\cite{He2016DeepRL} and VGG-small~\cite{Zhang2018LQNetsLQ} on CIFAR-10~\cite{krizhevsky2009learning}, and ResNet-18/34~\cite{He2016DeepRL} on ImageNet~\cite{russakovsky2015imagenet}. The experimental results show that ReCU achieves state-of-the-art performance, as well as faster training convergence even with the simple STE~\cite{Bengio2013EstimatingOP} as our weight gradient approximation.

To sum up, this paper makes the following contributions:

\begin{itemize}
  \item We explore the influence of ``dead weights'' showing that they can adversely affect the optimization of BNNs. To the best of our knowledge, this is the first work to analyze the ``dead weights'' in BNNs.
  
  \item We introduce a rectified clamp unit (ReCU) to revive the ``dead weights'' and then provide a rigorous  mathematical proof that a smaller quantization error can be derived using our ReCU.
  
  \item A mathematical analysis on why the weight standardization helps boost BNNs is provided, and the inherent contradiction between minimizing the quantization error and maximizing the information entropy in BNNs is revealed. 
 
  \item Extensive experiments demonstrate that ReCU not only leads to better performance over many state-of-the-arts~\cite{Ding2019RegularizingAD, Qin2020ForwardAB, Lin2020RotatedBN, Zhou2016DoReFaNetTL, Gong2019DifferentiableSQ, Yang2020SearchingFL, Rastegari2016XNORNetIC, Chen2020BinarizedNA, Wan2018TBNCN, liu2018bi, Gu2019ProjectionCN, Gu2019BayesianO1, Cai2017DeepLW, Lin2017TowardsAB}, but also results in faster training convergence.
\end{itemize}

\section{Related Work}

As a pioneering work, Courbariaux \emph{et al}.~\cite{courbariaux2016binarized} binarizes both weights and activations with the sign function. To overcome the almost everywhere zero gradient in the sign function, they considers the STE~\cite{Bengio2013EstimatingOP} as an approximation to enable the gradient to back propagate. However, the representational ability of BNNs is very limited in a binary space, leading to a significant drop in accuracy. To mitigate the accuracy gap between BNN and its full-precision counterpart, XNOR-Net~\cite{Rastegari2016XNORNetIC} introduces a scaling factor, which is obtained through the $\ell_{1}$-norm of the weights or activations, to reduce the quantization error. XNOR-Net++~\cite{Bulat2019XNORNetIB} fuses the two scaling factors for quantized weights and activations into one parameter, and makes it learnable via the standard backpropagation. The rotated binary neural network (RBNN)~\cite{Lin2020RotatedBN} takes into account the influence of the angular bias between the binarized weight vector and its full-precision version, and then devises a bi-rotation scheme with two rotation matrices for angle alignment, which reduces the quantization error.

Other works propose to boost the performance of BNNs by devising new gradient estimation functions or designing quantization-friendly network architectures. For example,~\cite{Gong2019DifferentiableSQ,Yang2019QuantizationN,Lin2020RotatedBN} design a continuous activation gradient that gradually approximates the sign function so as to replace the conventional STE~\cite{Bengio2013EstimatingOP}. Qin \emph{et al}.~\cite{Qin2020ForwardAB} proposed an error decay estimator to minimize the information loss of gradients during backpropagation. ABC-Net~\cite{Lin2017TowardsAB} utilizes more binary bases for weights and activations to strengthen the model performance. ReActNet~\cite{liu2020reactnet} constructs a strong baseline by adding parameter-free shortcuts on top of MobileNetV1~\cite{howard2017mobilenets} and achieves 69.4\% top-1 accuracy on ILSVRC-2012. Leng \emph{et al}.~\cite{Leng2018ExtremelyLB} modelled the BNN learning as a discretization-constrained optimization problem solved by the ADMM optimizer, so as to avoid the non-differentiable quantization. In~\cite{Yang2020SearchingFL}, an auxiliary probability matrix is made to search for the discrete quantized weights, implemented in a differentiable manner. 

\section{Background}

In this section, we briefly review the optimization of BNNs. Given a DNN, for ease of representation, we simply denote its per-layer real-valued weights as $\mathcal{W}^r$ and the inputs as $\mathcal{A}^r$. Then, the convolutional result can be expressed as
\begin{equation}
    Y = \mathcal{A}^r \otimes \mathcal{W}^r,
\end{equation}
where $\otimes$ represents the standard convolution. For simplicity, we omit the non-linear operations in this paper.

BNN aims to binarize each weight $w^r \in \mathcal{W}^r$ and each activation $a^r \in \mathcal{A}^r$ to $\{+1, -1\}$. Following XNOR-Net~\cite{Rastegari2016XNORNetIC}, the binarization can be achieved by the sign function,
\begin{equation}\label{binarization}
    x^b=\operatorname{sign}(x^r)=\left\{\begin{array}{l}
        +1, \text{ if } x^r \ge 0, \\
        -1, \text{ otherwise.}
    \end{array}\right. 
\end{equation}

To mitigate the large quantization error in binarizing a DNN, XNOR-Net~\cite{Rastegari2016XNORNetIC} further introduces two scaling factors for the weights $\mathcal{W}^b$ and activations $\mathcal{A}^b$, respectively. In this paper, following~\cite{Bulat2019XNORNetIB},  we simplify these two scaling factors as one parameter, denoted as $\alpha$. Then, the binary convolution operation can be formulated as
\begin{equation}\label{binary_conv}
    Y \approx (\mathcal{A}^b \circledast \mathcal{W}^b) \odot \alpha,
\end{equation}
where $\circledast$ represents the bit-wise operations including XNOR and POPCOUNT, and $\odot$ denotes the element-wise multiplication. Then, the quantization error in a BNN is defined as
\begin{equation}\label{quantization_error}
    \operatorname{QE} = \int_{-\infty}^{+\infty}f(w^r)\big(w^r - \alpha \operatorname{sign}(w^r)\big)^2\mathrm{d}w^r,
\end{equation}
where $f(w^r)$ is the probability density function of $w^r$.

To train a BNN, the forward convolution is achieved using the $\mathcal{W}^b$ and $\mathcal{A}^b$ binarized by Eq.\,(\ref{binarization}), while the real-valued $\mathcal{W}^r$ and $\mathcal{A}^r$ are updated during backpropagation. However, the gradient of the sign function is zero-valued almost everywhere, which is not suitable for optimization. Instead, we use the simple STE~\cite{Bengio2013EstimatingOP} in this paper to compute the approximate gradient of the loss \emph{w.r.t.} $w^r \in \mathcal{W}^r$, as
\begin{equation}\label{gradient_weight}
    \dfrac{\partial \mathscr{L}}{\partial w^r}=\dfrac{\partial \mathscr{L}}{\partial w^{b}} \cdot \dfrac{\partial w^{b}} {\partial w^r} \approx \dfrac{\partial \mathscr{L}}{\partial w^{b}},
\end{equation}
where $\mathscr{L}$ is the loss function.

As for the gradient \emph{w.r.t.} the activations, we consider the piece-wise polynomial function~\cite{liu2018bi} as follows
\begin{equation}\label{gradient_input}
    \dfrac{\partial \mathscr{L}}{\partial a^r}=\dfrac{\partial \mathscr{L}}{\partial a^{b}} \cdot \dfrac{\partial a^{b}}{\partial a^r}\approx \dfrac{\partial \mathscr{L}}{\partial a^{b}} \cdot \dfrac{\partial F(a^r)}{\partial a^r},
\end{equation}
where
\begin{equation}
    \dfrac{\partial F(a^r)}{\partial a^r}= \left\{
    \begin{array}{ll}
        2 + 2a^r, &\text{ if } -1 \le a^r < 0, \\
        2 - 2a^r, &\text{ if }\quad\, 0  \le a^r < 1, \\
        0, &\text{ otherwise. } 
    \end{array}\right.
\end{equation}

From now on, we drop the superscript ``$r$'' for real-valued weights for simplicity.


\section{Methodology}

\subsection{The Dead Weights in BNNs}\label{dead}

As pointed out in~\cite{Zhong2020TowardsLB,banner2019post}, the latent weights $\mathcal{W}$ of a quantized network roughly follow the zero-mean Laplace distribution due to their quantization in the forward propagation. As can be seen from Fig.\,\ref{dead_weight_illu}, most weights are gathered around the distribution peak (origin point), while many outliers fall into the two tails, far away from the peak.

We argue that these outliers adversely affect the training of a BNN and might be the potential reason for the slow convergence when training BNNs. Specifically, in real-valued networks, weights of different magnitudes make different contributions to the network performance; in other words, it is how far each weight is from the origin that matters. In BNNs, however, there is not much distinction between weights of different magnitudes if they have the same sign since only the signs are kept in the forward inference regardless of their magnitudes. Therefore, from the perspective of optimization, though the magnitudes of the weights are updated during the backpropagation by gradient descent, the chances of changing their signs are unequal. Intuitively, the signs of the weights around the distribution peak are easily changed, while it is the opposite for the outliers in the tails, which greatly limits the representational ability of BNNs and thus causes slow convergence in training. For this reason, we call these outliers ``dead weights'' in BNNs.

To solve this problem, in Sec.\,\ref{clamp}, we introduce our rectified clamp unit to revive these ``dead weights'' along with a rigorous proof that our clamp function leads to a smaller quantization error. In Sec.\,\ref{diversity}, we analyze why the weight standardization can boost the performance of BNNs, and reveal the inherent contradiction between minimizing the quantization error and maximizing the information entropy of the weights. Correspondingly, in Sec.\,\ref{experiment}, we introduce an adaptive exponential scheduler to identify the range of the ``dead weights'' in order to seek a balance between the quantization error and the information entropy.

\subsection{Rectified Clamp Unit}\label{clamp}

To solve the aforementioned problem, we propose ReCU, which aims to move the ``dead weights'' towards the distribution peak to increase the probability of changing their signs. Specifically, for each real-valued weight $w \in \mathcal{W}$, ReCU is formulated as
\begin{equation}
    \operatorname{ReCU}(w) = \operatorname{max}\Big(\operatorname{min}\big(w, Q_{(\tau)}\big), Q_{(1-\tau)}\Big),
    \label{eq:recu}
\end{equation}
where $Q_{(\tau)}$ and $Q_{(1-\tau)}$ respectively denote the $\tau$ quantile and $1-\tau$ quantile~\cite{zwillinger2002crc} of $\mathcal{W}$. With $0.5< \tau \le 1$, ReCU relocates $w$ to $Q_{(1-\tau)}$ if it is smaller than $Q_{(1-\tau)}$, and to $Q_{(\tau)}$ if it is larger than $Q_{(\tau)}$. In this way, the ``dead weights'' are revived. To measure the contribution of ReCU in the quantization process, we take into account the quantization error for analysis. In what follows, we show that the weights after applying ReCU can derive a smaller quantization error.

Earlier works~\cite{Zhong2020TowardsLB,banner2019post} have shown that the latent weights roughly follow the zero-mean Laplace distribution, \emph{i.e.}, $w \sim La(0, b)$, which implies $Q_{(\tau)}+Q_{(1-\tau)}=0$. Thus, we have
%

\begin{equation}
    \int_{-\infty}^{Q_{(\tau)}}\dfrac{1}{2b}\exp(-\dfrac{|w|}{b})\mathrm{d}w = \tau,
\end{equation}
which results in
\begin{equation}
    Q_{(\tau)}=-b\ln(2-2\tau).
\end{equation}

However, it is difficult to know the exact value of $b$. Luckily, we can obtain its approximation via the maximum likelihood estimation, represented as
\begin{equation}\label{estimation_b}
    \hat{b} = \operatorname{Mean}(|\mathcal{W}|),
\end{equation}
where $\operatorname{Mean}(|\cdot|)$ returns the mean of the absolute values of the inputs. Thus, $Q_{(\tau)}$ is a function of $\tau$.

After applying ReCU to $w$, the generalized probability density function of $w$ can be written as follows
\begin{equation}
    f(w) = \left\{\begin{array}{ll}
         \dfrac{1}{2b}\exp(\dfrac{-|w|}{b}), &\text{ if } |w|<Q_{(\tau)}, \\
         1-\tau, &\text{ if } |w|=Q_{(\tau)}, \\
         0, &\text{ otherwise. }
    \end{array} 
    \right.
\end{equation}

To obtain the quantization error, we first compute the scaling factor in Eq.\,(\ref{binary_conv}) using the Riemann-Stieltjes integral as
\begin{equation}\label{alpha}
    \begin{split}
        \alpha 
        =& \mathbb{E}(|\mathcal{\operatorname{ReCU}(W)}|) \\
        =& \int_{-Q_{(\tau)}}^{Q_{(\tau)}}|w|f(w)\mathrm{d}w+\sum_{|w|=Q_{(\tau)}}|w|f(w)\\
        =& \int_{0}^{Q_{(\tau)}}\dfrac{w}{b}\exp(-\dfrac{w}{b})\mathrm{d}w+2Q_{(\tau)}(1-\tau)\\
        =& b-(Q_{(\tau)}+b)\exp(-\dfrac{Q_{(\tau)}}{b}) + 2Q_{(\tau)}(1-\tau).
    \end{split}
\end{equation}

\begin{figure}[!t]
\begin{center}
\includegraphics[width=0.65\linewidth]{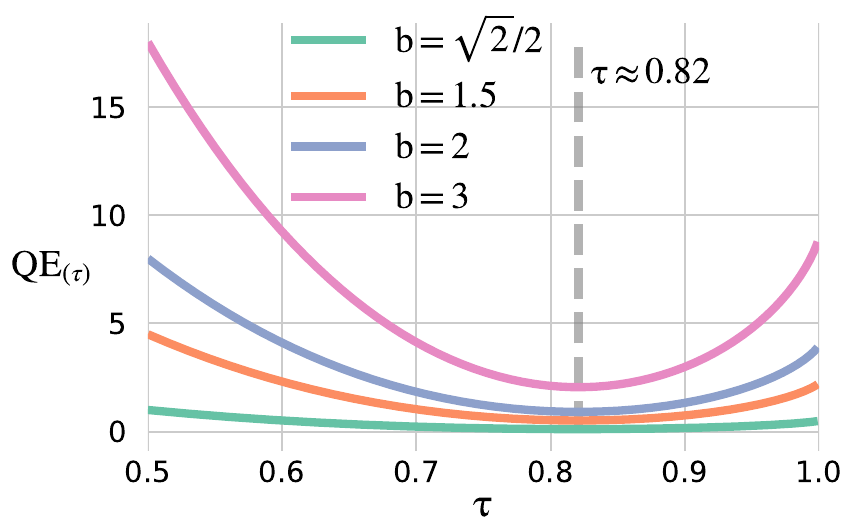}
\end{center}
\vspace{-1.5em}
   \caption{Visualization of the quantization error after applying ReCU. As can be seen, $\operatorname{QE}(\tau)$ is a convex function of $\tau$ and reaches the minimum when $\tau \approx 0.82$.}
\vspace{-1.5em}
\label{Fig:fig3}
\end{figure}

Obviously, $\alpha$ is a function of $\tau$ after replacing $b$ with the estimation $\hat{b}$ in Eq.\,(\ref{estimation_b}). With $Q_{(\tau)} + Q_{(1-\tau)} = 0$, the quantization error in Eq.\,(\ref{quantization_error}) after ReCU is derived as


\begin{equation}
    \begin{split}
    \operatorname{QE}(\tau) &= \int_{-\infty}^{+\infty}f(w)\big(w - \alpha \operatorname{sign}(w)\big)^2\mathrm{d}w
    \\& = \int_{-Q_{(\tau)}}^{Q_{(\tau)}}f(w)\big(w - \alpha \operatorname{sign}(w)\big)^2\mathrm{d}w
    \\& \quad + \sum_{| w | = Q_{(\tau)}}f(w)\big(w - \alpha \operatorname{sign}(w)\big)^2
    \\& = \int_{0}^{Q_{(\tau)}}\dfrac{1}{b}\exp(-\dfrac{w}{b})(w-\alpha)^{2}\mathrm{d}w
    \\& \quad + \sum_{| w | = Q_{(\tau)}}(1 - \tau)\big(w - \alpha \operatorname{sign}(w)\big)^2    
    \\&=(\alpha-b)^{2}\big(1 + \exp(-\dfrac{Q_{(\tau)}}{b})\big)+b^{2}
    \\& \quad -  \big((b+Q_{(\tau)})^{2}-2\alpha Q_{(\tau)}\big)\exp(-\dfrac{Q_{(\tau)}}{b})
    \\& \quad +2(1-\tau)(Q_{(\tau)}-\alpha)^{2}.
    \end{split}
    \label{eq:QE}
\end{equation}

Similarly, setting $b = \hat{b}$ makes $\operatorname{QE}(\tau)$ a function of $\tau$. From Eq.\,(\ref{eq:QE}), we have two observations: (1) As plotted in Fig.\,\ref{Fig:fig3}, $\operatorname{QE}(\tau)$ is a convex function when $0.5 < \tau \le 1$ and reaches the minimum when $\tau \approx 0.82$\footnote{A rigorous proof is provided in the Appendix.}. (2) When $\tau = 1$, Eq.\,(\ref{eq:QE}) degenerates to the normal quantization error as defined in Eq.\,(\ref{quantization_error}) where ReCU is not introduced.

However, we cannot keep $\tau = 0.82$ to pursue the least quantization error. In the next subsection, we analyze another important factor to the network performance, \emph{i.e.}, information entropy, which requires $\tau > 0.82$ to support good performance, and reveal the contradiction between minimizing the quantization error and maximizing the information entropy. Overall, we have the following inequality
\begin{equation}
    \operatorname{QE}(\tau) \le \operatorname{QE}(1) = \operatorname{QE},  \quad 0.82 \le \tau \le 1.
\end{equation}
That is, ReCU provides a smaller quantization error than $\operatorname{QE}$ when $0.82 \le \tau < 1$.

\subsection{Information Entropy of Weights}\label{diversity}
\label{section:4.3}
The information entropy of a random variable is the average level of uncertainty in the variable's possible outcomes, which is also used as a quantitative measure to reflect the weight diversity in BNNs~\cite{Lin2020RotatedBN, Qin2020ForwardAB, Raj2020UnderstandingLD, Liu2019CirculantBC}. Usually, the more diverse, the better the performance of a BNN. Given a probability density function $p(x)$ on domain $\mathcal{X}$, the information entropy is defined as
\begin{equation}
    H(p)=\mathbb{E}\big(-\ln(p(x))\big)=-\int_{\mathcal{X}}p(x)\ln\big(p(x)\big)\mathrm{d}x.
    \label{eq:Hp}
\end{equation}

Accordingly, the information entropy of $\mathcal{W}$ after applying ReCU can be computed by
\begin{equation}
    \begin{split}
        H(f)
        =&-\int_{-Q_{(\tau)}}^{Q_{(\tau)}}f(w)\ln\big(f(w)\big)\mathrm{d}w
        \\& -\sum_{|w|=Q_{(\tau)}}f(w)\ln\big(f(w)\big)\\
        =&-\int_{0}^{Q_{(\tau)}}\dfrac{1}{b}\exp(-\dfrac{w}{b})\ln\big(-\dfrac{1}{2b}\exp(-\dfrac{w}{b})\big)\mathrm{d}w
        \\& - 2(1-\tau)\ln(1-\tau) \\
        =& \; 2(\ln b+1)\tau + \ln\dfrac{2}{b}-1,
    \end{split}
\end{equation}
which is a function of $\tau$ by substituting $\hat{b}$ in Eq.\,(\ref{estimation_b}) for $b$.

\begin{figure}[!t]
\begin{center}
\includegraphics[width=0.7\linewidth]{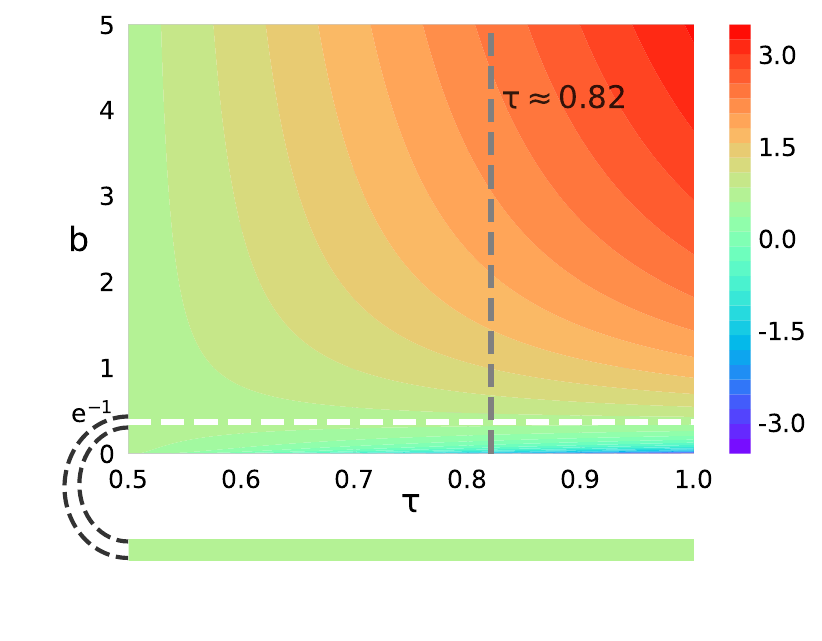}
\end{center}
\vspace{-1em}
   \caption{The information entropy of $\mathcal{W}$ \emph{w.r.t.} $\tau$ and $b$. The dotted white line indicates a fixed value of the information entropy when $b = e^{-1}$. (Best viewed in color)}
\vspace{-1em}
\label{Fig:fig4}
\end{figure}

For ease of the following analysis, we visualize the information entropy \emph{w.r.t.} varying values of $b$ and $\tau$ in Fig.\,\ref{Fig:fig4}. Then, we have

\textit{Case 1}: $b = e^{-1}$. In this situation, the information entropy $H(f)$ is fixed to $\ln2$ (the dotted white line in Fig.\,\ref{Fig:fig4}).

\textit{Case 2}: $b < e^{-1}$. $H(f)$ is a monotonically decreasing function of $\tau$.

\textit{Case 3}: $b > e^{-1}$. $H(f)$ is a monotonically increasing function of $\tau$.

Recall that $b$ is estimated by the mean of the absolute values of $\mathcal{W}$ in Eq.\,(\ref{estimation_b}). We have experimentally observed that in the cases of $b \le e^{-1}$, the information entropy ($\le \ln 2$) is too small to enable good performance (see Sec.\,\ref{effect_bstar}). Thus, a larger $b$ should be derived to overcome this problem. However, in practice, the weights $\mathcal{W}$ gradually become sparse during network training due to the widely-used $\ell_p$-norm regularization in modern neural networks, making the information entropy uncontrollable, which inevitably brings a loss of diversity.

Thus, it is necessary to maintain $b$ at a relatively high value (\textit{Case 3}) in a controllable manner to retain the information entropy. The previous work~\cite{Qin2020ForwardAB} maximizes the information entropy by centralizing and standardizing the weights in each forward propagation as follows
\begin{equation}
    \mathcal{W}^\prime = \dfrac{\mathcal{W}-\mathbb{E}(\mathcal{W})}{\sigma (\mathcal{W})},
    \label{eq:w_hat}
\end{equation}
where $\sigma(\cdot)$ denotes the standard deviation. 
In what

However, we experimentally find that it is the standardization, but not the centralization, that contributes to the performance improvement. The reason comes from the fact that $\mathbb{E}(\mathcal{W}) \approx 0$ in most cases \cite{He2019SimultaneouslyOW, Qin2020ForwardAB}. This motivates us to generalize Eq.\,(\ref{eq:w_hat}) by simply standardizing the weights $\mathcal{W}$ as
\begin{equation}
    \mathcal{W}^\prime = \frac{\mathcal{W}}{K},
\end{equation}
where $K > 0$ is a given constant. Then, the mean of the absolute values of the weights after standardization is
\begin{equation}
    b^\prime  = \operatorname{Mean}(|\mathcal{W}^\prime|)=\dfrac{b}{K}.
\end{equation}

It is easy to see that $\sigma(\mathcal{W})=\sqrt{2}b$ due to the Laplace distribution. Therefore, by setting $K = \sigma(\mathcal{W})$ as in~\cite{Qin2020ForwardAB}, $b^\prime$ becomes
\begin{equation}\label{information_irnet}
    b^\prime = \dfrac{b}{\sqrt{2}b} = \dfrac{\sqrt{2}}{2}>e^{-1},
\end{equation}
which increases the information entropy and explains why dividing $\mathcal{W}$ by the standard deviation can result in better performance when training BNNs~\cite{Qin2020ForwardAB}. To the best of our knowledge, this is the first time that a mathematical explanation is provided. Nevertheless, according to Fig.\,\ref{Fig:fig4}, the information entropy can be increased with a larger $b$. Thus, we further define $K = \sigma(\mathcal{W}) / (\sqrt{2}b^{\star})$ where $b^{\star}$ is a pre-defined constant, and the standardized weights $\mathcal{W}^\prime$ become
\begin{equation}
    \mathcal{W}^\prime = \dfrac{\mathcal{W}}{\sigma(\mathcal{W})/ (\sqrt{2}b^\star)}.
    \label{eq:w_tilde}
\end{equation}
It is easy to see that
\begin{equation}
    b^\prime = b^\star.
\end{equation}

The innovation behind this analysis lies in that our standardization transforms the uncontrolled information entropy to an adjustable one by manually setting $b^\star$ based on the premise of $b^\star > e^{-1}$, and therefore generalizes the information gain of Eq.\,(\ref{eq:w_hat}) by~\cite{Qin2020ForwardAB}. Thus, by standardizing the weights using Eq.\,(\ref{eq:w_tilde}) before applying ReCU, the information entropy can be increased in the learning of a BNN. 

Nevertheless, the information increase from enlarging $b$ is still very limited (see Fig.\,\ref{Fig:fig4}). In contrast, the increase of $\tau$ leads to more information gain, with an unexpected increase in quantization error when $\tau > 0.82$, as analyzed in Sec.\,\ref{clamp}. Thus, there exists an inherent contradiction between minimizing the quantization error and maximizing the information entropy in BNNs. To the best of our knowledge, we are the first to find this contradiction. In Sec.\,\ref{effect_tau}, we propose an exponential scheduler for adapting $\tau$ along the network training, so as to seek a balance between the information entropy and the quantization error.

\textbf{Training Procedures}. Given a DNN with its per-layer real-valued weights $\mathcal{W}$ and the inputs $\mathcal{A}$, in the forward propagation, we first standardize $\mathcal{W}$ and revive the ``dead weights'' using Eq.\,(\ref{eq:w_tilde}) and ReCU of Eq.\,(\ref{eq:recu}), respectively. Then, we compute the scaling factor $\alpha$ using Eq.\,(\ref{alpha}), and binarize the inputs and the revived weights using the sign function of Eq.\,(\ref{binarization}). Finally, we complete the binary convolution using Eq.\,(\ref{binary_conv}) for the forward propagation. During backpropagation, we derive the gradients \emph{w.r.t.} $\mathcal{W}$ and $\mathcal{A}$ using Eq.\,(\ref{gradient_weight}) and Eq.\,(\ref{gradient_input}), respectively, and update $\mathcal{W}$ using the stochastic gradient descent (SGD) described in Sec.\,\ref{implementation}.

\section{Experiments}\label{experiment}

In this section, we evaluate ReCU on the two widely-adopted CIFAR-10~\cite{krizhevsky2009learning} and ILSVRC-2012~\cite{russakovsky2015imagenet} datasets, and then compare it to several state-of-the-art methods~\cite{Qin2020ForwardAB, liu2018bi, Lin2020RotatedBN, Yang2020SearchingFL, Gong2019DifferentiableSQ}.  

\subsection{Implementation}\label{implementation}




\textbf{Network Structures.} On CIFAR-10, we evaluate ReCU with ResNet-18/20~\cite{He2016DeepRL} and VGG-Small~\cite{Zhang2018LQNetsLQ}. Following the compared methods, we binarize all convolutional and fully-connected layers except the first and the last ones. For ResNet-18/20, we adopt the double skip connections as proposed in~\cite{liu2018bi} for fair comparison.

On ILSVRC-2012, we choose to binarize ResNet-18/34. Following~\cite{Bethge2019BackTS}, the downsampling layers are not quantized. Similarly, the double skip connections~\cite{liu2018bi} are added.


\textbf{Training Details.} Our network is trained from scratch without depending on a pre-trained model. For all experiments, we use SGD for optimization with a momentum of 0.9 and the weight-decay is set to 5e-4. The initial learning rate is 0.1 and then adjusted by the cosine scheduler~\cite{Loshchilov2016SGDRSG}. We follow the data augmentation strategies in~\cite{He2016DeepRL} which include random crops and horizontal flips.  

\subsection{Ablation Study}

In this section, we discuss the hyperparameter settings of ReCU, including $b^\star$ and $\tau$. Recall that $b^\star$ affects the information entropy, while $\tau$ affects both the quantization error and the information entropy. Each experiment is run three times and we report the mean top-1 accuracy (mean $\pm$ std) of ResNet-18 (64-64-128-256) for parametric analyses.

\begin{table}[!t]
\centering
\renewcommand\arraystretch{1.1}
\caption{Top-1 accuracy of ResNet-18 \emph{w.r.t}. different values of $\tau$ on CIFAR-100.\label{table:tau}}
\begin{minipage}{\textwidth}
\begin{minipage}{0.25\textwidth}
\centering
\begin{tabular}{cc}
\hline
$\tau$  &   mean $\pm$ std (\%)\\ \hline
1.00       &   67.55$~\pm$~0.07   \\[-2pt] \hline
0.98       &   68.10$~\pm$~0.09   \\[-2pt] \hline
0.96       &   68.06$~\pm$~0.13   \\[-2pt] \hline
0.94  &  68.29$~\pm$~0.06   \\[-2pt] \hline
\textbf{0.92}       &   \textbf{68.47$~\pm$~0.09}   \\[-2pt] \hline
\end{tabular}
\end{minipage}
\begin{minipage}{0.25\textwidth}
\begin{tabular}{cc}
\hline
$\tau$  &   mean $\pm$ std (\%)\\ \hline
0.90    &   68.18$~\pm$~0.09   \\[-2pt] \hline
0.85    &   67.39$~\pm$~0.26   \\[-2pt] \hline
0.82    &   66.50$~\pm$~0.15   \\[-2pt] \hline
0.80    &   63.56$~\pm$~0.26   \\[-2pt] \hline
0.78    &   59.09$~\pm$~0.20   \\[-2pt] \hline
\end{tabular}
\end{minipage}
\end{minipage}
\end{table}

\subsubsection{Effect of $\tau$ for ReCU}\label{effect_tau}

In Sec.\,\ref{clamp}, we demonstrate that the quantization error with ReCU is a convex function of $\tau$ and becomes the minimum when $\tau \approx 0.82$, while the information entropy is a monotonically increasing function of $\tau$ if $b^{\star} > e^{-1}$. Thus, a balance needs to be reached between the quantization error and the information entropy. To this end, following~\cite{Qin2020ForwardAB} \big(Eq.\,(\ref{eq:w_hat})\big), we set $b^{\star} = \sqrt{2}/2$ for our analyses.

We first consider setting $\tau$ to a fixed value for the whole training process. As shown in Tab.\,\ref{table:tau}, when $\tau = 0.92$, the network reaches the best performance. It is worth noting that a significant drop in accuracy occurs when $\tau < 0.82$. This is understandable since it suffers both a large quantization error and small information entropy. Another observation is that ReCU does not obtain the best accuracy when $\tau = 0.82$. This is because though the quantization error reaches the minimum when $\tau = 0.82$ as shown in Fig.\,\ref{Fig:fig3}, the small information entropy cannot support a good performance. In summary, when $0.85 \le \tau \le 1.00$, we can seek a balance between the quantization error and the information entropy.

Despite its good performance when using a fixed value of $\tau$, we find that ReCU increases the variance of the performance when $0.85 \le \tau \le 0.94$ while keeping it stable when $0.96 \le \tau \le 1.00$ as shown in Tab.\,\ref{table:tau}. To solve this, we further propose an exponential scheduler for adapting $\tau$ along the network training. Our motivation lies in that $\tau$ should start with a value falling within $[0.85, 0.94]$ to pursue a good accuracy, and then gradually go to the interval $[0.96, 1.00]$ to stabilize the variance of performance. Based on this, given an initial $\tau_{s}$ and an end threshold $\tau_{e}$, $\tau_i$ at the $i$-th training epoch is calculated as follows

\begin{equation}\label{adaptive}
    \tau_{i} = \dfrac{\tau_{e}-\tau_{s}}{e-1}e^{i/I}+\dfrac{e \cdot \tau_{s}-\tau_{e}}{e-1},
\end{equation}
where $I$ denotes the total number of training epochs.

Tab.\,\ref{table:tau_exp} shows that ReCU obtains better performance of $68.69\%$ with $\tau_s = 0.85$ and $\tau_e = 0.99$. Besides, it can well overcome the large variance with a fixed $\tau$.

\begin{table}[!t]
\renewcommand\arraystretch{1.1}
\caption{Top-1 accuracy of ResNet-18 \emph{w.r.t}. $\tau$ calculated by our exponential scheduler on CIFAR-100.}
\begin{minipage}{\textwidth}
\begin{minipage}{0.24\textwidth}
\setlength{\tabcolsep}{0.2em}
\begin{tabular}{ccccccc}
\hline
$\tau_{s} / \tau_{e}$ & mean~$\pm$~std (\%) \\ \hline
0.80 / 1.00 & 68.37~$\pm$~0.16\\[-2pt] \hline
0.85 / 1.00 & 68.55~$\pm$~0.11\\[-2pt] \hline
0.90 / 1.00 & 68.50~$\pm$~0.10\\[-2pt] \hline
\end{tabular}
\end{minipage}
\begin{minipage}{0.24\textwidth}
\setlength{\tabcolsep}{0.2em}
\begin{tabular}{ccccccc}
\hline
$\tau_{s} / \tau_{e}$ & mean~$\pm$~std (\%) \\ \hline
0.80 / 0.99 & 68.44~$\pm$~0.15\\[-2pt] \hline
\textbf{0.85 / 0.99} & \textbf{68.69~$\pm$~0.13}\\[-2pt] \hline
0.90 / 0.99 & 68.61~$\pm$~0.17\\[-2pt] \hline
\end{tabular}
\end{minipage}
\end{minipage}
\label{table:tau_exp}
\end{table}

\subsubsection{Effect of $b^\star$ for Weight Standardization}\label{effect_bstar}

Tab.\,\ref{table:bstar} displays the results \emph{w.r.t.} different values of $b^\star$. we use Eq.\,(\ref{adaptive}) for adapting $\tau$ with $\tau_{s}=0.85$ and $\tau_{e}=0.99$. The experiments are conducted under three settings for a comprehensive analysis, including training the BNN without our standardization, $b^\star = 0.2 <e^{-1}$, and $b^\star > e^{-1}$.

As can be observed, without our standardization, the binarized ResNet-18 shows a poor top-1 accuracy of $66.13\%$. For a detailed analysis, during network training, the $\ell_p$-norm regularization in the current neural network sparsifies the network parameters, which reduces the information entropy as discussed in Sec.\,\ref{diversity}.

With our standardization in hand, the information entropy can be manually controlled by adjusting $b^\star$. In Tab.\,\ref{table:bstar}, with a small $b^\star = 0.2 < e^{-1}$, despite the better performance of $68.10\%$, the improvement is limited. As discussed in Sec.\,\ref{diversity}, the information entropy is still too small to enable good performance when $b^\star \le e^{-1}$.

We further set $b^\star > e^{-1}$. As can be seen, the network reaches a maximum mean top-1 accuracy of 69.02\% when $b^\star = 2$, a significant improvement over the model without our standardization. We also observe that as $b^\star$ continues to increase, the performance starts to remain stable, which supports our claim in Sec.\,\ref{diversity} that the improvement from enlarging $b^\star$ is limited.

\begin{table}[!t]
\renewcommand\arraystretch{1.1}
\caption{Top-1 accuracy of ResNet-18 \emph{w.r.t}. different values of $b^\star$ for weight standardization on CIFAR-100. ``w/o'' denotes binarization without our standardization.}
\centering
\begin{minipage}{\textwidth}
\begin{minipage}{0.25\textwidth}
\centering
\begin{tabular}{cccccccc}
\hline
$b^\star$    & mean~$\pm$~std (\%) \\[-2pt] \hline
w/o          & 66.13~$\pm$~0.21 \\[-2pt] \hline
0.2          & 68.10~$\pm$~0.17 \\[-2pt] \hline
$\sqrt{2}/2$ & 68.69~$\pm$~0.13 \\[-2pt] \hline
1            & 68.82~$\pm$~0.11 \\[-2pt] \hline
\end{tabular}
\end{minipage}
\begin{minipage}{0.25\textwidth}
\begin{tabular}{cccccccc}
\hline
$b^\star$ & mean~$\pm$~std (\%) \\[-2pt] \hline
\textbf{2} & \textbf{69.02~$\pm$~0.07}\\[-2pt] \hline
3          & 68.98~$\pm$~0.10\\[-2pt] \hline
4          & 68.80~$\pm$~0.15\\[-2pt] \hline
5          & 68.48~$\pm$~0.13\\[-2pt] \hline
\end{tabular}
\end{minipage}
\end{minipage}
\label{table:bstar}
\end{table}

\begin{table}[!t]
\renewcommand\arraystretch{1.1}
\caption{Top-1 accuracy of ResNet-18 \emph{w.r.t}. different training epochs on CIFAR-100.}
\centering
\begin{tabular}{ccccccc}
\hline
Training epochs & 100   & 300   & 600     \\ \hline
Vanilla         & 52.1 & 59.6 & 62.0      \\[-2pt] \hline 
Ours            & 66.3 & 68.2 & 69.1      \\[-2pt] \hline
\end{tabular}
\label{table:convergence}
\end{table}

\begin{figure}[!t]
\begin{center}
\includegraphics[width=0.7\linewidth]{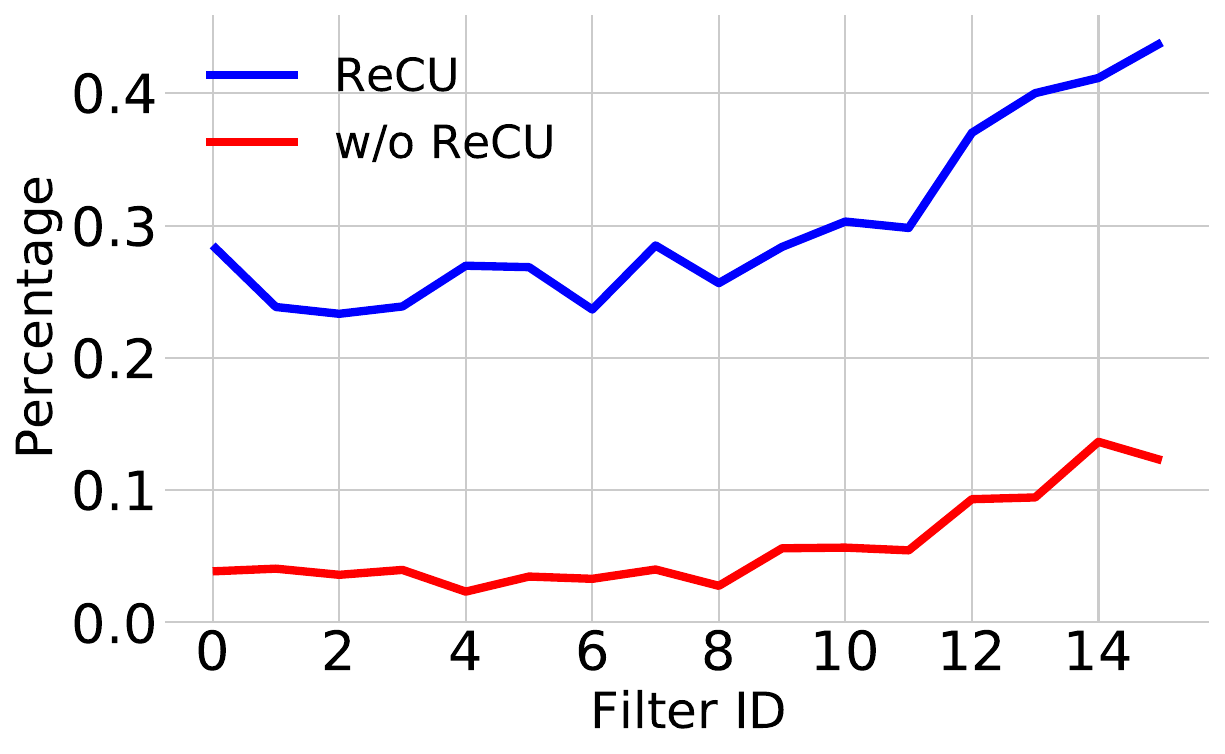}
\end{center}
   \caption{Flipping percentage of ``dead weights'' with and without our ReCU (ResNet-18 on CIFAR-100).}
\label{append}
\end{figure}

\subsection{Evidence on Reviving Dead Weights}

We compare the sign difference (flipping percentage) of ``dead weights'' within 20\% the largest magnitudes at points of half and final training epochs in Fig.\,\ref{append}. As can be seen, less than 10\% of ``dead weights'' are updated without ReCU. In contrast, about 13\% to 35\% are updated with ReCU. Thus, our ReCU greatly revives ``dead weights''.

\subsection{Training Convergence}\label{convergence}
As discussed in Sec.\,\ref{dead}, the ``dead weights'' introduces an obstacle to the training convergence of BNNs. In Tab.\ref{table:convergence}, we show the effectiveness of ReCU in overcoming this problem. As seen, ReCU achieves 66.3\% top-1 accuracy with only 100 training epochs, while the vanilla BNN obtains 62.0\% even when it is trained for 600 epochs. 

\subsection{Comparison with SOTA Methods}

To quantitatively evaluate the effectiveness of the proposed ReCU, we conduct extensive experiments on CIFAR-10~\cite{krizhevsky2009learning} and ImageNet~\cite{russakovsky2015imagenet}. We also compare it with a number of state-of-the-art methods to demonstrate the advantages of ReCU in boosting the performance of BNNs. In the following experiments, we use Eq.\,(\ref{adaptive}) for adapting $\tau$ with $\tau_{s}=0.85$ and $\tau_{e}=0.99$. Besides, $b^\star$ is set to 2.

\begin{table}[t]
\renewcommand\arraystretch{1.1}
\caption{Performance comparison with the state-of-the-arts on CIFAR-10. W/A denotes the bit length of the weights and activations. FP is short for full precision. \label{table:cifar}}
\centering
\begin{tabular}{cccc}
\toprule
Network                    & Method     & W/A   & Top-1 \\ \toprule
\multirow{5}{*}{ResNet-18} & FP         & 32/32 & 94.8\% \\ 
& RAD~\cite{Ding2019RegularizingAD}     & 1/1   & 90.5\% \\  
& IR-Net~\cite{Qin2020ForwardAB}        & 1/1   & 91.5\%  \\ 
& RBNN~\cite{Lin2020RotatedBN}          & 1/1   & 92.2\%  \\
& \textbf{ReCU (Ours)}                  & 1/1   & \textbf{92.8}\%  \\
\toprule
\multirow{6}{*}{ResNet-20} & FP         & 32/32 & 92.1\% \\
& DoReFa~\cite{Zhou2016DoReFaNetTL}     & 1/1   & 79.3\% \\
& DSQ~\cite{Gong2019DifferentiableSQ}   & 1/1   & 84.1\% \\ 
& SLB~\cite{Yang2020SearchingFL}        & 1/1   & 85.5\% \\
& IR-Net~\cite{Qin2020ForwardAB}        & 1/1   & 86.5\% \\
& \textbf{ReCU (Ours)}                  & 1/1   & \textbf{87.4}\% \\
\toprule
\multirow{10}{*}{VGG-small} & FP        & 32/32 & 94.1\% \\
& XNOR-Net \cite{Rastegari2016XNORNetIC}& 1/1   & 89.8\% \\
& BNN~\cite{courbariaux2016binarized}   & 1/1   & 89.9\% \\ 
& DoReFa~\cite{Zhou2016DoReFaNetTL}     & 1/1   & 90.2\% \\
& IR-Net~\cite{Qin2020ForwardAB}        & 1/1   & 90.4\% \\
& RBNN~\cite{Lin2020RotatedBN}          & 1/1   & 91.3\% \\
& DSQ~\cite{Gong2019DifferentiableSQ}   & 1/1   & 91.7\% \\
& SLB~\cite{Yang2020SearchingFL}        & 1/1   & 92.0\% \\
& \textbf{ReCU(Ours)}                   & 1/1   & \textbf{92.2}\% \\
\toprule
\end{tabular}
\end{table}

\subsubsection{CIFAR-10}

For ResNet-18, we compare ReCU with RAD~\cite{Ding2019RegularizingAD}, RBNN~\cite{Lin2020RotatedBN} and IR-Net~\cite{Qin2020ForwardAB}. For ResNet-20, the compared methods include SLB~\cite{Yang2020SearchingFL}, DSQ~\cite{Gong2019DifferentiableSQ}, DoReFa~\cite{Zhou2016DoReFaNetTL}, and IR-Net~\cite{Qin2020ForwardAB}. For VGG-Small, we compare ReCU with XNOR-Net~\cite{Rastegari2016XNORNetIC}, DoReFa~\cite{Zhou2016DoReFaNetTL}, BNN~\cite{courbariaux2016binarized}, SLB~\cite{Yang2020SearchingFL}, IR-Net~\cite{Qin2020ForwardAB}, RAD~\cite{Ding2019RegularizingAD}, DSQ~\cite{Gong2019DifferentiableSQ}, and RBNN~\cite{Lin2020RotatedBN}. The experimental results are shown in Tab.\,\ref{table:cifar}. As can be observed, ReCU shows the best performance in all the networks. Specifically, with ResNet-18, ReCU obtains 0.6\% performance increase over the recent RBNN. Also, it yields 0.9\% performance gain over IR-Net in binarizing ResNet-20. Lastly, it retains a top-1 accuracy of 92.2\% when binarizing VGG-small, which is better than the search-based SLB result of 92.0\%.

\begin{table}[t]
\renewcommand\arraystretch{1.1}
\caption{Performance comparison with the state-of-the-arts on ImageNet. W/A denotes the bit length of the weights and activations. FP is short for full precision. ReCU* means using the same network and training setting as ReActNet~\cite{liu2020reactnet}.\label{table:imagenet}}
\centering
\setlength{\tabcolsep}{0.38em}
\begin{tabular}{ccccc}
\toprule
Network                   & Method & W/A & Top-1 & Top-5 \\ \toprule
\multirow{13}{*}{ResNet-18} & FP        &32/32  & 69.6\%    & 89.2\% \\
& BNN~\cite{courbariaux2016binarized}   & 1/1   & 42.2\%    & 67.1\% \\
& XNOR-Net~\cite{Rastegari2016XNORNetIC}& 1/1   & 51.2\%    & 73.2\% \\
& TBN~\cite{Wan2018TBNCN}               & 1/2   & 55.6\%    & 79.0\% \\
& Bi-Real Net~\cite{liu2018bi}          & 1/1   & 56.4\%    & 79.5\% \\
& PCNN~\cite{Gu2019ProjectionCN}        & 1/1   & 57.3\%    & 80.0\% \\
& IR-Net~\cite{Qin2020ForwardAB}        & 1/1   & 58.1\%    & 80.0\% \\
& DoReFa~\cite{Zhou2016DoReFaNetTL}     & 1/4   & 59.2\%    & -      \\
& BONN~\cite{Gu2019BayesianO1}          & 1/1   & 59.3\%    & 81.6\% \\
& HWGQ~\cite{Cai2017DeepLW}             & 1/2   & 59.6\%    & 82.2\% \\
& RBNN~\cite{Lin2020RotatedBN}          & 1/1   & 59.9\%    & 81.9\% \\
& \textbf{ReCU(Ours)}                   & 1/1   & \textbf{61.0}\% &\textbf{82.6}\% \\
& ReActNet~\cite{liu2020reactnet}       & 1/1   & 65.9\%    & -     \\
& \textbf{ReCU*(Ours)}                  & 1/1   & \textbf{66.4}\% &\textbf{86.5}\% \\
\toprule
\multirow{6}{*}{ResNet-34} & FP         &32/32 & 73.3\%     & 91.3\% \\
& ABC-Net~\cite{Lin2017TowardsAB}       & 1/1  & 52.4\%     & 76.5\%\\
& Bi-Real Net~\cite{liu2018bi}          & 1/1  & 62.2\%     & 83.9\%\\ 
& IR-Net \cite{Qin2020ForwardAB}        & 1/1  & 62.9\%     & 84.1\%\\
& RBNN~\cite{Lin2020RotatedBN}          & 1/1  & 63.1\%     & 84.4\%\\ 
& \textbf{ReCU(Ours)}                   & 1/1  & \textbf{65.1}\%     &\textbf{85.8}\%\\
\toprule
\end{tabular}
\end{table}

\subsubsection{ImageNet}

Tab.\,\ref{table:imagenet} displays the performance comparison in binarizing ResNet-18/34. For ResNet-18, we compare ReCU with BNN~\cite{courbariaux2016binarized}, Bi-Real Net~\cite{liu2018bi}, XNOR-Net~\cite{Rastegari2016XNORNetIC}, SLB~\cite{Yang2020SearchingFL}, DoReFa~\cite{Zhou2016DoReFaNetTL}, IR-Net~\cite{Qin2020ForwardAB}, RBNN~\cite{Lin2020RotatedBN}, PCNN~\cite{Gu2019ProjectionCN} and BONN~\cite{Gu2019BayesianO1}. For ResNet-34, ABC-Net~\cite{Lin2017TowardsAB}, Bi-Real Net~\cite{liu2018bi}, IR-Net~\cite{Qin2020ForwardAB} and RBNN~\cite{Lin2020RotatedBN} are compared. From Tab.\,\ref{table:imagenet}, we can see that ReCU takes the leading place in both the top-1 and top-5 accuracies. Specifically, it obtains a better performance of 61.0\% in top-1 and 82.6\% in top-5 compared to RBNN's 59.9\% top-1 accuracy and 81.9\% top-5 accuracy with ResNet-18. We also compare with ReActNet\footnote{\url{https://github.com/liuzechun/ReActNet\#models}}~\cite{liu2020reactnet}, and under the same network and training setting, we obtain 0.5\% performance improvement. The advantage in ResNet-34 is even more obvious where ReCU obtains 2.0\% and 1.4\% performance improvements in top-1 and top-5 accuracy, respectively. 

Thus, Tab.\,\ref{table:cifar} and Tab.\,\ref{table:imagenet} verify the correctness of exploring the ``dead weights'' and the effectiveness of ReCU in reviving them, which greatly benefits the performance of BNNs.

\section{Conclusion}

In this paper, we present a novel rectified clamp unit (ReCU) to revive the ``dead weights'' when training BNNs. We first analyze how the ``dead weights'' block the optimization of BNNs and slow down the training convergence. Then, ReCU is applied to increase the probability of changing the signs of the ``dead weights'' on the premise of a rigorous proof that ReCU can lead to a smaller quantization error. Besides, we analyze why the weight standardization can increase the information entropy of the weights, and thus benefit the BNN performance. The inherent contradiction between minimizing the quantization error and maximizing the information entropy is revealed for the first time. Correspondingly, an adaptive exponential scheduler is proposed to seek a balance between the quantization error and the information entropy. Experimental results demonstrate that, by reviving the ``dead weights'', ReCU leads to not only faster training convergence but also state-of-the-art performance.

\section{Acknowledge}
This work is supported by the National Science Fund for Distinguished Young Scholars (No.62025603), the National Natural Science Foundation of China (No.U1705262, No. 62072386, No. 62072387, No. 62072389, No. 62002305,  
No.61772443, No.61802324 and No.61702136), Guangdong Basic and Applied Basic Research Foundation (No.2019B1515120049) and the Fundamental Research Funds for the central universities (No. 20720200077, No. 20720200090 and No. 20720200091).

{\small
\bibliographystyle{ieee_fullname}
\bibliography{egbib}
}

\section*{Appendix \label{appendix}}

\textbf{Convexity and Minimum of $\operatorname{QE}(\tau)$}

We first revisit the formulations for the $\tau$ quantile $Q_{(\tau)}$ and scaling factor $\alpha$ in the following

\begin{equation}\label{supple_quantile}
    Q_{(\tau)}=-b\ln(2-2\tau). \tag{10}
\end{equation}

\begin{equation}\label{supple_alpha}
    \begin{split}
        \alpha 
        =& b-(Q_{(\tau)}+b)\exp(-\dfrac{Q_{(\tau)}}{b}) 
        + 2Q_{(\tau)}(1-\tau).
    \end{split} \tag{13}
\end{equation}

Combining Eq.\,(\ref{supple_quantile}) and Eq.\,(\ref{supple_alpha}), we can rewrite $\alpha$ as
\begin{equation}\label{supple_alpha_a1}
    \alpha = b(2\tau - 1). \tag{A1}
\end{equation}

Recall that the quantization error under our framework is given as

\begin{equation}\label{supple_quan_err}
\begin{split}
    \operatorname{QE}(\tau) &=(\alpha-b)^{2}\big(1 + \exp(-\dfrac{Q_{(\tau)}}{b})\big)+b^{2}
    \\& \quad -  \big((b+Q_{(\tau)})^{2}-2\alpha Q_{(\tau)}\big)\exp(-\dfrac{Q_{(\tau)}}{b})
    \\& \quad +2(1-\tau)(Q_{(\tau)}-\alpha)^{2}.
\end{split} \tag{14}
\end{equation}

By putting Eq.\,(\ref{supple_quantile}) and Eq.\,(\ref{supple_alpha_a1}) into Eq.\,(\ref{supple_quan_err}), we reformulate $\operatorname{QE}(\tau)$ as

\begin{equation}\label{supple_quan_err_sim}
\begin{split}
     &\operatorname{QE}(\tau) 
     \\&
     = b^{2}\big(-16\tau^{3}+44\tau^{2}-40\tau-4(\tau-1)\ln(2-2\tau)+13\big). \\
\end{split} \tag{A2}
\end{equation}

According to Eq.\,(\ref{supple_quan_err_sim}), the derivative of $\operatorname{QE}(\tau)$ \emph{w.r.t.} $\tau$ can be derived as
\begin{equation}
\label{pqe}
\begin{split}
\dfrac{\partial \operatorname{QE}(\tau)}{\partial \tau} &= -4b^{2}(12\tau^{2}-22\tau+\ln(2-2\tau)+11)
\\&=4b^{2}G(\tau),
\end{split}\tag{A3}
\end{equation}
where 
\begin{equation}
    G(\tau)=-12\tau^{2}+22\tau-\ln(2-2\tau)-11. \tag{A4}
\end{equation}

Note that $b$ is estimated via the maximum likelihood estimation as
\begin{equation}\label{supple_estimation_b}
    \hat{b} = \operatorname{Mean}(|\mathcal{W}|), \tag{11}
\end{equation}
which indicates $b \neq 0$. We can know that
\begin{equation}\label{supple_partial_qe_tau_A5}
    \dfrac{\partial \operatorname{QE}(\tau)}{\partial \tau}=0 \iff G(\tau)=0.\tag{A5}
\end{equation}

Thus, the extreme value of $\operatorname{QE}(\tau)$ is irrelevant to $b$. Further, we yield the derivative of $G(\tau)$ \emph{w.r.t.} $\tau$ as
\begin{equation}\label{supple_partial_t_tau_A6}
    \dfrac{\partial G(\tau)}{\partial \tau}=-24\tau+\dfrac{1}{1-\tau}+22. \tag{A6}
\end{equation}

From Eq.\,(\ref{supple_partial_t_tau_A6}), it is easy to know that $\dfrac{\partial G(\tau)}{\partial \tau}>0$ if $\tau \le 1$. Therefore, $G(\tau)$ is monotonically increasing when $\tau \le 1$. By solving $G(\tau)=0$, we have $\tau \approx 0.82$. That means when $0.82< \tau \le 1$, $G(\tau)>0$, while when $0.5<x<0.82$, $G(\tau)<0$. That is to say, $\dfrac{\partial QE(\tau)}{\partial \tau}<0$ when $0.5<\tau < 0.82$, and $\dfrac{\partial \operatorname{QE}(\tau)}{\partial \tau}>0 \text{ when } 0.82<\tau \le 1$. Thus $\operatorname{QE}(\tau)$ is a convex function \emph{w.r.t.} $\tau \in (0.5,1]$ and reaches the minimum when $\tau \approx 0.82$, which completes the proof. $\hfill\blacksquare$

\end{document}